\documentclass{article}

\usepackage{arxiv}

\usepackage{textpos}
\usepackage[numbers]{natbib} 
\usepackage{tabularx,longtable,multirow,subfigure,caption}
\usepackage{fancyhdr} 
\usepackage{url} 
\usepackage[english]{babel}
\usepackage[linesnumbered]{algorithm2e}
\usepackage{amsmath}
\usepackage{amsfonts} 
\usepackage{graphicx}
\usepackage{dsfont}
\usepackage{epstopdf} 
\usepackage{array}
\usepackage{latexsym}
\usepackage{hyperref}
\usepackage{titlesec}
\usepackage{booktabs}
\titleformat{\chapter}[display]  
{\normalfont\huge\bfseries}{\chaptertitlename\ \thechapter}{20pt}{\LARGE}  
\titlespacing{\chapter}{0pt}{0pt}{0pt} 

\hypersetup{pdftitle={},
  pdfsubject={}, 
  pdfauthor={},
  pdfkeywords={}, 
  pdfstartview=FitH,
  pdfpagemode={UseOutlines},
  bookmarksnumbered=true, bookmarksopen=true, colorlinks,
    citecolor=black,%
    filecolor=black,%
    linkcolor=black,%
    urlcolor=black}

\newcommand{\probP}{\text{I\kern-0.15em P}}

\date{Written 4 January 2024, revised 6 January 2024}

\title{Evolution-Bootstrapped Simulation: \\ Artificial or Human Intelligence: Which Came First?}

\author{Paul A. Bilokon}

\begin{document}

\maketitle

\textbf{Paul A. Bilokon} is the CEO and Founder of Thalesians in Canary Wharf, London E14 5AB, and a visiting professor at Imperial College London.\\
\underline{paul.bilokon@imperial.ac.uk}\\
Department of Mathematics, Imperial College London, South Kensington, London SW7 2BX\\\\

\begin{abstract}

Humans have created artificial intelligence (AI), not the other way around. This statement is deceptively obvious. In this note, we decided to challenge this statement as a small, lighthearted Gedankenexperiment. We ask a simple question: in a world driven by evolution by natural selection, would neural networks or humans be likely to evolve first? We compare the Solomonoff--Kolmogorov--Chaitin complexity of the two and find neural networks (even LLMs) to be significantly simpler than humans. Further, we claim that it is unnecessary for any complex human-made equipment to exist for there to be neural networks. Neural networks may have evolved as naturally occurring objects before humans did as a form of chemical reaction-based or enzyme-based computation. Now that we know that neural networks can pass the Turing test and suspect that they may be capable of superintelligence, we ask whether the natural evolution of neural networks could lead from pure evolution by natural selection to what we call evolution-bootstrapped simulation. The evolution of neural networks does not involve irreducible complexity; would easily allow irreducible complexity to exist in the evolution-bootstrapped simulation; is a falsifiable scientific hypothesis; and is independent of / orthogonal to the issue of intelligent design.

\end{abstract}


Humans have created artificial intelligence~(AI), not the other way around. This statement is deceptively obvious. In this note, we decided to challenge it as a small Gedankenexperiment. This note is intended as lighthearted, so should be taken with scepticism.

The first question is whether AI was created or discovered. Gravity certainly existed (and made massive objects fall to the ground) long before Newton formulated his law of universal gravitation~\cite{Newton1687} or Aristotle observed that objects immersed in a medium tend to fall at speeds proportional to their weight~\cite{Aristotle1934}. So there is little doubt that Aristotle, Newton, and others discovered certain aspects of gravity --- but they did not \emph{create} gravity.

With AI the situation is somewhat less clear, because AI research involves both science and engineering --- AI systems must be implemented before they can be said to exist. All known AI systems have been implemented by humans, therefore we assume that for AI to exist, humans must exist first.

The scientific consensus is that humans exist as a result of evolution by natural selection~\cite{Darwin2009, Hurst2009, Dawkins2010}. One of the arguments levelled against evolution by natural selection is the so-called \emph{irreducible complexity~(IC)}~\cite{Behe1996}: the argument that certain biological systems with multiple interacting parts would not function if one of the parts were removed, so supposedly could not have evolved by successive small modifications from earlier, less complex systems through natural selection, which would need all intermediate precursor systems to have been fully functional.

IC has been engaged by proponents of \emph{intelligent design~(ID)} or \emph{creationists}, who continue the line of thought started by IC by adding that the only alternative explanation is a ``purposeful arrangement of parts,'' i.e. design by an intelligent agent~\cite{Davis1993}.

The expression ``irreducible complexity'' was introduced by Behe in his 1996 book \textit{Darwin's black box}~\cite{Behe1996}, Behe being himself an advocate of ID. Behe and others have suggested a number of biological features that they believed to be irreducibly complex: the blood clotting cascade; the eye; flagella; cilium motion; and the bombardier beetle's defense mechanism.

The mainstream scientific community has been sceptical about IC and its variations. The arguments against irreducible complexity went along several lines. For example, potentially viable evolutionary pathways for allegedly irreducibly complex systems such as blood clotting, the immune system, and the flagellum --- the three examples proposed by Behe, --- and the mousetrap, have been proposed~\cite{Aird2003, Bottaro2006, Pallen2006, McDonald2011}. Some have argued that irreducible complexity may not exist in nature, and the examples proposed by Behe and others may not represent irreducible complexity, but can be explained in terms of simpler precursors. Gerhart and Kirschner have presented the \emph{theory of facilitated variation}~\cite{Gerhart2007}, which describes how certain mutations can cause apparent irreducible complexity.

Others have pointed out that precursors of complex systems, when they are not useful in themselves, may be useful to perform other, unrelated functions. A 2006 article in Nature demonstrates intermediate states leading toward the development of the ear in a Devonian fish about 360 million years ago~\cite{Daeschler2006}.

Irreducible complexity can be seen as equivalent to an ``uncrossable valley'' in a fitness landscape~\cite{Trotter2014}. A number of mathematical models of evolution have explored the circumstances under which such valleys can be crossed~\cite{Weissman2009, Weissman2010-1, Covert2013, Trotter2014}. For example, the protein T-urg13, which is responsible for the cytpolasmic male sterility of waxy corn and is due to a completely new gene is claimed to be irreducibly complex in Dembski's book \textit{No Free Lunch}~\cite{Dembski2006} has been shown to have evolved~\cite{Chen2017}.

Other critics of irreducible complexity, such as Coyne~\cite{Coyne2009} and Scott~\cite{Scott2009}, have argued that the concept of irreducible complexity and, more generally, intelligent design, is not falsifiable and, therefore, unscientific.

Thus most of the arguments levelled against irreducible complexity tend to:
\begin{enumerate}
\item challenge the very existence of irreducible complexity in nature;
\item propose alternative methods for crossing the seemingly ``uncrossable valleys'';
\item reject the argument as not falsifiable and, therefore, unscientific.
\end{enumerate}

Let us consider the following question: which came first, the neural network or the human? At first sight, this question appears absurd: certainly neural networks were developed by human pioneers~\cite{McCulloch1943, Rosenblatt1958, Rumelhart1986, LeCun2015, Goodfellow2016}.

Now, let us ask a different question: assuming a world with evolution by natural selection, which is more likely to have evolved first, a neural network or a human? Is it at all possible that a neural network could predate a human?

While evolution can lead to both increases and decreases in complexity~\cite{Saunders1976}, genome complexity has generally increased since the beginning of the life on earth~\cite{Sharov2006}. Proteins tend to become more hydrophobic over time~\cite{Wilson2017}, and to have their hydrophobic amino acids more interspersed along the primary sequence~\cite{Foy2019}. Increases in body size over time are sometimes seen in what is known as Cope's rule~\cite{Heim2015}.

Which is more complex, a neural network or a human? It depends on how you define complexity. One definition that can apply to both the neural network and the human is the \emph{Solomonoff--Kolmogorov--Chaitin complexity}~\cite{Solomonoff1997, Kolmogorov1998, Chaitin1969}: the complexity of an object is defined as the length of a shortest computer program (in a predetermined programming language) that produces the object as output. The Solomonoff--Kolmogorov--Chaitin complexity of a neural network is negligible compared to the complexity of a human because of the neural network's self-similarity: the layers of a neural network have a similar or identical structure and are themselves composed of identical neurons or units. Depending on the type of the neural network, these units can be more or less complex, but for most neural networks, such as the feedforward neural networks, the units are very simple. The entire structure can be specified very succinctly using loops or recursion in modern programming languages.

Moreover, there is no irreducible complexity in the presently known types of neural networks. A perceptron is already a functioning computational device. A two-layer feedforward neural network can be grown naturally from a single perceptron through self-similarity (again, obtaining a functioning computational device), and the process can be repeated to obtain deep neural networks. This evolution through self-similarity is consistent with Mandelbrot's ``fractal geometry of nature''~\cite{Mandelbrot1999} and resembles the growth of crystals and chain polymerisation. From the standpoint of Solomonoff--Kolmogorov--Chaitin complexity, neural networks, even large language models (LLMs), are very simple objects.

One would argue that the neural network cannot exist in isolation, that it need computer equipment to exist, so it couldn't evolve. However, in 1994, Adleman described the experimental use of DNA as a computational system~\cite{Adleman1994}. He solved a seven-node instance of the Hamiltonian Graph problem, an NP-complete problem similar to the travelling salesman problem. Later, chemical reaction networks~\cite{Shah2020, Chen2013, Srinivas2017, Soloveichik2010} and enzyme-based DNA computers~\cite{Shapiro2012} have been shown to be Turing-equivalent. In other words, there may be no need for humans or computers to exist for there to be functioning neural networks in this world; a naturally occurring chemical reaction-based or enzyme-based DNA computation would suffice.

One would then argue that life is required for DNA to exist. However, consider viroids~\cite{Navarro2021}, small single-stranded, circular RNAs that are infectious pathogens. Unlike viruses, they have no protein coating. A recent metatranscriptomics study~\cite{Lee2023} suggests that the host diversity of viroids and other viroid-like elements is broader than previously thought, and that it would not be limited to plants, encompassing even the prokaryotes.

Viruses, prions, and viroids are non-living organisms that require a living cellular host in order to reproduce. They cannot do it on their own. These parasites may be just a string of RNA, as in a viroid, or a length of DNA enclosed in a protein shell, as in a virus. However, given a replication mechanism, there is no reason why a naturally occurring viroid (for example) could not exist, a viroid that could implement a neural net, perhaps of increasing complexity. This is just one way in which naturally occurring neural networks could have evolved independently of --- and potentially before --- humans. These neural networks wouldn't require the existence of humans --- to create or discover them, nor is irreducible complexity involved in their evolution.

This is not the stuff of science fiction. At least, \emph{human-made} nanoparticle-based computing architectures for nanoparticle neural networks already exist~\cite{Kim2020}, even if we haven't (yet?) found naturally occurring structures of this kind. The idea of evolving neural networks and, more broadly, \emph{evolutionary connectivism}, is not new~\cite{Chalmers1991, Stanley2002, Ma2022}.

One objection would be that the training of a neural network~\cite{Rumelhart1986} is a relatively complex procedure. Could irreducible complexity be lurking somewhere in this process? Unlikely. When viewed through the prism of \emph{equilibrium propagation}~\cite{Scellier2017}, the training of a neural network could be interpreted as an energy-based model, reminiscent of energy minimization processes ubiquitous in the physical chemistry of polymers (see, for instance, \cite{Lin2023}). Equilibrium propagation does not need a special computation or circuit for the second phase, where errors are explicitly propagated. Because the objective function is defined in terms of local perturbations, the second phase of equilibrium propagation corresponds to only nudging the prediction (fixed point or stationary distribution) toward a configuration that reduces prediction error. This discovery makes it more plausible that a similar mechanism could be implemented by brains, since leaky integrator neural computation performs both inference and error backpropagation in this model.

We now know that neural networks are capable of intelligence~\cite{Koubaa2023} --- and potentially, superintelligence~\cite{Bostrom2017}, --- pass the Turing test~\cite{Turing1950}, and can run complex simulations. What would happen if, at some point, evolving neural networks started simulating evolution, perhaps imperfectly, allowing ``uncrossable valleys'' to be crossed through pure ``creativity'' (in other words, leaps across the space of internal representations, along the lines of generative AI)~\cite{Vikhar2016}? This would enable truly irreducibly complex structures to emerge as computational phenomena in a simulated environment (along the lines of Bostrom~\cite{Bostrom2003}\footnote{One key difference from Bostrom's framework is that EBS removes the need for there to be a parent or host civilization, unless you consider deep intelligence such a civilization.}) running within a system driven purely by natural selection and involving no irreducible complexity. This hypothesis of \emph{evolution-bootstrapped simulation~(EBS)} is falsifiable, and therefore scientific. As for ID, this argument seeks neither to confirm, nor to deny it: the hypothesis of evolution-bootstrapped simulation is orthogonal to ID.

If EBS were true and has \emph{already} happened instead of being a \emph{future} event (in principle, there is no reason why multiple EBSs could not be nested), then AI would not be that artificial, and it would probably make sense to call it \emph{deep intelligence} rather than artificial intelligence. Artificial intelligence would then be deep intelligence recreating itself (with human help) within the simulated system. Perhaps the alchemical symbol of Ouroboros has been misinterpreted: instead of eating itself, Ouroboros is emitting itself out. This, of course, raises a number of issues. Would an EBS world be subject to apparent limitations of deep intelligence, such as \emph{catastrophic inference}~\cite{McCloskey1989}? What would such limitations (if they are real) correspond to in the physics of the EBS world?

The existence of a physical evolutionary system, perhaps along the lines of~\cite{Lin2023} but self-organizing along the lines of energy minimization~\cite{Scellier2017}, would provide some evidence towards EBS. As to whether EBS has already happened and/or is yet to happen, various mathematical, physical, computational and other anomalies (e.g., widespread violations of the central limit theorem~\cite{Cuzzocrea2021}) would provide some evidence towards a transition towards EBS in the past. Interestingly, whereas true irreducible complexity weakens the argument for evolution by natural selection, it strengthens the argument for EBS.

\newcommand{\etalchar}[1]{$^{#1}$}

\end{document}